# Towards the Fully Automatic Merging of Lexical Resources: a Step Forward

**Muntsa Padró, Núria Bel and Silvia Necşulescu**
Universitat Pompeu Fabra
Roc Boronat, 138, ES-08018-Barcelona
E-mail: muntsa.padro@upf.edu, nuria.bel@upf.edu, silvia.necsulescu@upf.edu

**Abstract**

This article reports on the results of the research done towards the fully automatically merging of lexical resources. Our main goal is to show the generality of the proposed approach, which have been previously applied to merge Spanish Subcategorization Frames lexica. In this work we extend and apply the same technique to perform the merging of morphosyntactic lexica encoded in LMF. The experiments showed that the technique is general enough to obtain good results in these two different tasks which is an important step towards performing the merging of lexical resources fully automatically.

## 1. Introduction

The automatic production, updating, tuning and maintenance of Language Resources for Natural Language Processing is currently being considered as one of the most promising areas of advancement for the full deployment of Language Technologies. The reason is that these resources that describe, in one way or another, the characteristics of a particular language are necessary for Language Technologies to work for that particular language.

Although the re-use of existing resources such as WordNet (Fellbaum, 1998) in different applications has been a well known and successful case, it is not very frequent. The different technology or application requirements, or even the ignorance about the existence of other resources, has provoked the proliferation of different, unrelated resources that, if merged, could constitute a richer repository of information augmenting the number of potential uses. This is especially important for under-resourced languages, which normally suffer from the lack of broad coverage resources.

Several attempts of resource merging have been addressed and reported in the literature. Hughes et al. (1995) report on merging corpora with more than one annotation scheme. Ide and Bunt (2010) also report on the use of a common layer based on a graph representation for the merging of different annotated corpora. Teufel (1995) and Chan and Wu (1999) were concerned with the merging of several source lexica for part-of-speech tagging. The merging of more complex lexica has been addressed by Crouch and King (2005) who produced a Unified Lexicon with lexical entries for verbs based on their syntactic subcategorization in combination with their meaning, as described by WordNet (Fellbaum, 1998), Cyc (Lenat, 1995) and VerbNet (Kipper et al., 2000).

Despite the undeniable achievements of the research just mentioned, most of it reports the need for a significant amount of human intervention to extract the information of existing resources and to map it into a format in which both lexica can be compared. The cost of this manual effort might explain the lack of more merging attempts. Therefore, any cost reduction would have a high impact in the actual re-use of resources.

In this context, a proposal such as the Lexical Markup Framework, LMF (Francopoulo et al. 2008) is also an attempt to standardize the format of computational lexica as a way to reduce the complexities of merging lexica. However, LMF (ISO-24613:2008) "does not specify the structures, data constraints, and vocabularies to be used in the design of specific electronic lexical resources". Therefore, the merging of two LMF lexica is certainly easier, but only if both also share the structure and vocabularies, if not, mapping has still to be done by hand. Our aim is to work towards the full automatization of the whole merging process. This constituted the main challenge of the research reported in Bel et al. (2011), where a method to perform the merging of two different lexical resources fully automatically was proposed. They applied the proposed method to the particular case of merging two very different subcategorization frame (SCF) lexica for Spanish obtaining encouraging results.

The aim of the research we present here was to assess to what extent the actual merging of information contained in different LMF lexica can be done automatically, following the mentioned method, in two cases: when the lexica to be merged share structure and labels, and when they do not. Besides, our second goal was to prove the generality of the approach, i.e. if it could be applied to different types of lexical resources.

Therefore, for this work we applied the method presented in Bel et al. (2011) to merge different Spanish morphosyntactic dictionaries. A first experiment tackled the merging of a number of dictionaries of the same family: Apertium monolingual lexica developed independently for different bilingual MT modules. A second experiment merged the results of the first experiments with the Spanish morphosyntactic FreeLing lexicon. All the lexica were already in the LMF format, although Apertium and FreeLing have different structure and tagset. In addition, note that these morphosyntactic lexica contain very different information than SCF lexica of the first experiments, and that what we present here can be considered a further proof of the good performance and generality of the proposed automatic merging method.

The current results have shown that the availability of the lexica to be merged in a common format such as LMF indeed alleviates the problem of merging. In our experiment with different Apertium lexica it was possible to merge three different monolingual morphosyntactic lexica with the method proposed as to achieve a larger resource. We have also obtained good results in the merging of different tag set based lexica.

## 2. Methodology

Basically, the merging of lexica has two well defined steps (Crouch and King, 2005):

1. Mapping step: because information about the same phenomenon can be expressed differently, the information in the existing resources has to be extracted and mapped into a common format.
2. Combination Step: once the information in both lexica is encoded in the same way, this information from both lexica is mechanically compared and combined to create the new resource.

Thus, our goal is to carry out the two steps of the merging process in a fully automatic way. This is to perform both mapping and combination steps without any human supervision.

In this section, we will first describe the lexica we wanted to merge, after we will discuss the problems of the combination step, which is simpler and motivates the mapping, which we will explain later.

### 2.1. The lexica

We have worked with Apertium lexica. Apertium (Armentano-Oller et al., 2007) is an open source rule-based MT system. In this framework, bilingual MT systems are developed independently (and by different people), and this also holds for the lexica for the same language that belong to different bilingual systems. These lexica that share format and tags can differ in the number of entries and the particular encoding of particular entries. For our experiments we merged three Spanish monolingual lexica coming from the Catalan-Spanish with 39,072 entries, English-Spanish with 30,490 entries and French-Spanish with 21,408 entries. In table 1 we further describe details of these lexica. Thus, we found numerous cases of common entries, missing entries in some of them, and also some phenomena related to homography (i.e. the same lemma with different morphological paradigm) as it is the case of *contador* that in one lexicon appears as the machine ('meter'), only

```
Apertium source:
    tenebrosísimo:tenebroso<adj><sup><m><sg>
Apertium LMF:
    <LexicalEntry id="id20588-s">
    <feat att="partOfSpeech" val="adj"/>
    <Lemma>
        <feat att="writtenForm" val="tenebroso"/>
    </Lemma>
    <WordForm>
        <feat att="writtenForm" val="tenebrosísimo"/>
        <feat att="type" val="sup"/>
        <feat att="gender" val="m"/>
        <feat att="number" val="sg"/>
    </WordForm>

FreeLing source:
    tenebroso tenebroso AQ0MS0
FreeLing LMF:
    <LexicalEntry>
    <feat att="partOfSpeech" val="adjectiveQualifier"/>
    <Lemma>
        <feat att="writtenForm" val="tenebroso"/>
    </Lemma>
    <WordForm>
        <feat att="writtenForm" val="tenebroso"/>
        <feat att="grade" val="-"/>
        <feat                 att="grammaticalGender"
        val="masculine"/>
        <feat                 att="grammaticalNumber"
        val="singular"/>
        <feat att="function" val="-"/>
    </WordForm>
```

Figure 1. Source and LMF versions for the adjective "tenebroso" ('gloomy') in Apertium and FreeLing lexica

masculine forms, and in other as the person with both feminine and masculine forms.

FreeLing morphosyntactic lexicon is used for morphological analysis and PoS disambiguation modules of the FreeLing NLP suite (Padró et al., 2010). It uses an adapted version of the EAGLES tag set (Leech and Wilson, 1999). The lexica, originally in the format shown in 1 as source, were converted into LMF in the framework of the METANET4U[1] project (the converted lexica are available in META-SHARE repository) but without changing the tag set labels. Although semantically very close (they both are describing morphosytactic data), main differences between the Apertium and FreeLing tag sets are in the way the information is encoded. For instance adjectives in FreeLing encode 'grade' and 'function', while in Apertium the grade was converted into a "type". The spelling of the name of attributes and values also vary in the source and in the converted files. Note that the conversion into LMF was done independently for each lexicon and followed the information supplied by the available documentation where the semantics of the tags was explained. The order of the features is maintained as well as the number of features.

---

[1] An EU PSP-CIP funded project whose aim is to make available and accessible updated and standardized language resources. The META-SHARE repository will be used for the distribution of such resources. www.meta-share.eu

## 2.2. Combination of lexica using graph unification

Necsulescu et al. (2011) and Bel et al. (2011) proposed to perform the combination step using graph unification (Kay, 1979). This single operation which is based on set union of compatible feature values, makes it possible to validate the common information, exclude the inconsistent one and to add, if desired, the unique information that each lexicon contained for building a richer resource. For graph unification in our experiments, we used the NLTK unification mechanism (Bird, 2006).

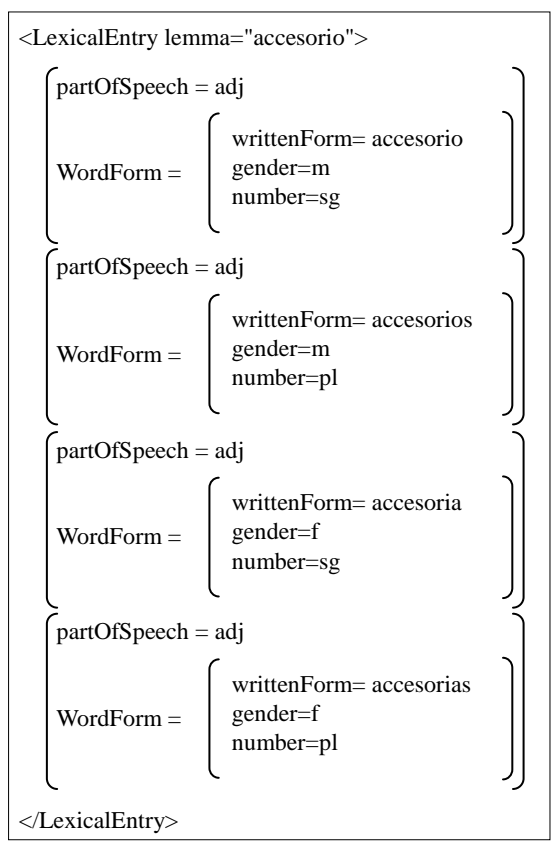

Figure 2. An Apertium LMF entry represented as a feature structure for graph unification.

In order to use graph unification, the LMF lexica had to be represented as feature structures (graphs, see figure 2 for a sample). Because LMF lexica already identified attributes and values in a structured way, this step was straightforward. Note that in converting LMF into feature structures, a lexical entry contains all its <WordForm> as present in the original lexicon together with the part of speech., while the lemma information is encoded as a special "feature" outside the feature structure in order to guide the unification process. This process is carried out by the system along the following steps:

1) For each lemma that is common to both lexica, it gathers all lexical entries with that lemma in both lexica (cases of homography are taken into account).

2) For the set of entries got in (1), it tries to unify every entry in one lexicon with all the entries in the other lexicon. This step implies checking unification for all feature structures included in the entries.

3) When having a successful unification, create an entry in the resulting lexicon. Unification operation will deliver as feature structures in the resulting entry those that resulting from the common information and also those present in one entry but not in the other.

4) When a lexical entry does not unify with anyone of the other lexicon, it creates an entry in the resulting lexicon as well, because it is considered to contain unique information.

5) For those lemmas that only are in one of the lexica, it creates a lexical entry in the resulting lexicon.

In order to be able to inspect the results, information about the operation that originated the entries in the resulting lexicon is registered in a log file.

## 2.3. Semantic Preserving Mapping

The proposal to avoid manual intervention when converting two lexica into a common format with a blind, semantic preserving method (Bel et al., 2011) departs from the idea of Chan and Wu (1999) of comparing information contained in common entries of different lexica and looking for significant equivalences in terms of consistent repetition. The basic requirement for this automatic mapping is to have a number of common entries encoded in the two lexica to be compared. Chan and Wu (1999) were working only with single part-of-speech tags, but the lexica we address here handle more complex and structured information, which has to be identified as units by the algorithm. In order to avoid the necessity of defining the significant pieces of information to be mapped by hand, Bel et al. (2011) proposed a method to first automatically identify such pieces ("minimal units") in each lexicon and secondly, to automatically learn the correspondence of such pieces between the two lexica. Their results showed that it is possible to assess that a piece of the code in lexicon A corresponds to a piece of code in lexicon B since a significant number of different lexical entries hold the same correspondence. Then, when a correspondence is found, the relevant piece in A is substituted by the piece in B, performing the conversion into the target format to allow for comparison and, eventually, merging as explained in section 2.2. Note that the task is defined in terms of automatically learning correspondences among both, labels and structure since both may differ across lexica. For example, in FreeLing the verb tense and mood are encoded in two different attributes (e.g. mood=subjunctive, tense=present), while Apertium encodes both tense and mood in a sole attribute (e.g.

tense=prs).

| Lexicon | Lexical Entries | Av. Word Forms per entry | Lexical Entries per PoS | | | | |
|---|---|---|---|---|---|---|---|
| | | | Nouns | Verbs | Adjectives | Adverbs | Proper nouns |
| Apertium | | | | | | | |
| Apertium ca-es | 39,072 | 7.35 | 16,054 | 4,074 | 5,883 | 4,369 | 8,293 |
| Apertium en-es | 30,490 | 6.41 | 11,296 | 2,702 | 4,135 | 1,675 | 10,084 |
| Apertium fr-es | 21,408 | 6.78 | 7,575 | 2,122 | 2,283 | 729 | 8,274 |
| **Aperium unified** (all) | 60,444 | 6.14 | 19,824 | 5,127 | 7,312 | 5,340 | 21,917 |
| FreeLing | | | | | | | |
| FreeLing | 76,318 | 8.76 | 49,519 | 7,658 | 18,473 | 169 | 0 |
| Apertium and Freeling | | | | | | | |
| **Apertium and FreeLing unified** ( mapping to FreeLing) | 112,621 | 7.03 | 54,830 | 8,970 | 20,162 | 5,406 | 21,917 |

Table 1: Original and unified lexicon sizes

The algorithm used in this work to learn the mapping between two lexica is basically the same used by Bel et al. (2011) although two changes were introduced in order to gain in generality. The main difference is due to the fact that in the first experiments with SCF lexica no attribute had an open list of values (for instance, the value of the attribute for 'writtenForm' does not have a closed number of possible values). We have made the algorithm more general, able to deal with a larger number of possible resource types by adding a different treatment for open and closed feature value types. The identification and special treatment of open values is made fully automatically and affect the step of finding units and learning the correspondence between lexica.

The identification of the open values is now the algorithm first step. By counting the different values in the lexicon, the system decides a feature value to be open when a relative large number of values are encountered. Open values are substituted with a variable in order to find the repetitions that are learnt as a pattern.

The other difference is that, because of the LMF source, we can work from the beginning with a feature structure version of the lexica, while in Bel et al. (2011) they worked with the source formats. Therefore, our algorithm splits each feature structure into feature-value pairs and looks for the elements that always occur together in order to identify the "minimal units" of information. This step is necessary in order to gain in generality when learning correspondences. Note that the probability of finding significant correspondences of larger units is lower. For instance, the system must learn that in FreeLing, *tense* and *mood* features always occur together, and that they both correspond to the information that is a value of the feature *tense* in Apertium.

In order to learn such mapping, for each possible pair of minimal units that are a potential mapping rule, the system measures the similarity in terms of the lemmas that contain a member of the pair in the corresponding lexica. That is, the list of lemmas that contain each minimal unit is represented as a binary vector and the Jaccard distance measure is used to compute similarity between vectors[2] (as Chan and Wu, 1999). The system chooses as correspondences those that maximize similarity, i.e., those with a larger number of lemmas that contain the minimal units to be mapped. In case that there is more than one correspondence, all are considered possible mappings.

Once the corresponding units have been identified, a new feature structure is created substituting units in lexicon A with the corresponding units of lexicon B. This operation results in a lexicon A encoded with the tagset of lexicon B. Now, both lexica can be compared and merged as explained in section 2.2. Note that the mapping should preserve the semantic of the feature value pair. Furthermore, this procedure also identifies the differences between the two lexica when no mapping for a particular minimal unit is found. This information can be later used for creating patches that systematically carry out corrective actions: direct transformation, deletion of feature structures, etc.

### 3. Experiments and Results

Our experiments were the following. We first merged the three Apertium lexica, and we evaluated the success of the combination step. For these three lexica, no mapping was required because they all use the same tagset. Once this merged lexicon was created, it was mapped and merged with the FreeLing lexicon. The results of the merging are presented in table 1.

---

[2] If the minimal unit is a sibling of an open valued feature, the elements in the vector are the values of this feature instead of the lemmas. E.g. "gender=m", is a sibling of "writtenForm", so the vector used will contain the values of "writtenForm".

From the results of merging all Apertium lexica, it is noticeable that the resulting Apertium lexicon has two times the entries (in average) of the source lexica, and that the part of speech that supplied more entries was proper noun. One can explain this if takes into account the independent development of the lexica and that each one probably took different reference test corpora. For the other parts of speech, there is a general increase of number of entries.

As for the merging with FreeLing lexicon experiment, in order to validate the results, both conversion senses were tested giving similar results. We will only comment on the Apertium into FreeLing as we have only closely inspected that experiment. From the data in table 1, we can see that again proper nouns but also adverbs are the main source of new entries. Because FreeLing did not include proper nouns, all the Apertium ones are added. Adverbs are also a major source of new elements, which can be explained because FreeLing handles derivate adverbs (adjective with the *–mente* suffix) differently to Apertium.

In what follows, we present separately the results of the two different steps, mapping and merging, for the Apertium into FreeLing lexica experiment. Also, concrete examples of the different cases are discussed.

In the mapping experiment from Apertium into FreeLing 127 and 152 minimal units were automatically identified respectively. The found mapping correspondences between them are shown in table 2.

| # possible mappings | #units |
|---|---|
| 0 | 19 |
| 1 | 99 |
| 2 | 8 |
| 3 | 1 |
| Total | 127 |

Table 2: Number of units that receive a concrete number of correspondences (mappings)

Note that mapping correspondences are learnt only if enough examples are seen. A threshold mechanism over the similarity measures controls the selection of the mapping rules to be applied. The most common cases were learnt satisfactorily, and the mapping of units with the lowest frequency had different results. For instance, the mapping of Apertium "type=sup" for superlative adjectives was not found to be correlated with the FreeLing "grade=superlative", mainly due to the little number of examples in FreeLing. On the other hand, Apertium lexicon contained only two examples of "future of subjunctive" but in FreeLing lexicon all verbs do have these forms and the system correctly learnt the mapping. There were also incorrect mappings, which, however, affected only few cases which could be traced back after the inspection of the inferred mapping rules.

Finally, there were some cases where no correspondence was found and a manual inspection of these cases confirmed that, indeed, they should not have a mapping. For example, there were some PoS tags in Apertium that had no correspondence in FreeLing: *proper noun* and *acronym*. The merging mechanism was the responsible of adding the entries with these tags to the resulting lexica.

As we said before, the lexical entries in the resulting lexicon may have three different origins: from unification of an entry in lexicon A and in lexicon B; from entries that did not unify although having the same lemma, and from entries whose lemma was not in one of the lexica. In the following tables a summary of the results of the different unification results are given.

| PoS | # LE | PoS | # LE |
|---|---|---|---|
| adjectiveQualifier | 5,206 | interjection | 13 |
| adpositionPreposition | 24 | nounCommon | 14,147 |
| adverbGeneral | 112 | pronoun | 4 |
| conjunctionCoordinated | 4 | pronounExclamative | 8 |
| conjunctionSubordinated | 8 | pronounIndefinite | 12 |
| determinantExclamative | 4 | pronounRelative | 9 |
| determinantIndefinite | 12 | | |

Table 3: Number of entries with the same information in lexicon A and in lexicon B per categories

| PoS | # LE | PoS | # LE |
|---|---|---|---|
| adjectiveQualifier | 561 | determinantIndefinite | 4 |
| adpositionPreposition | 0 | interjection | 2 |
| adverbGeneral | 11 | nounCommon | 792 |
| conjunctionCoordinated | 1 | pronounDemonstrative | 3 |
| conjunctionSubordinated | 1 | pronounExclamative | 2 |
| determinantIndefinite | 4 | pronounIndefinite | 1 |
| determinantExclamative | 0 | pronounPersonal | 4 |
| verbAuxiliary | 1 | pronounPossesive | 7 |
| verbMain | 3,929 | pronounRelative | 1 |

Table 4: Entries that gained information with the unification per categories

| PoS | #LE | PoS | #LE |
|---|---|---|---|
| adjectiveOrdinal | 4 | num | 11 |
| adjectiveQualifier | 1,138 | preadv | 11 |
| adpositionPreposition | 1 | pronoun | 2 |
| adverbGeneral | 41 | pronounDemonstrative | 2 |
| adverbNegative | 1 | pronounExclamative | 2 |
| cnjsub | 1 | pronounIndefinite | 33 |
| conjunctionCoordinated | 5 | pronounPersonal | 13 |
| conjunctionSubordinated | 13 | pronounPossessive | 3 |
| determinantArticle | 1 | pronounRelative | 3 |
| determinantDemonstrative | 5 | np | 5 |
| determinantExclamative | 1 | punctuation | 1 |
| determinantIndefinite | 28 | vbmod | 2 |
| determinantPossessive | 2 | verbAuxiliary | 1 |
| interjection | 51 | verbMain | 8 |
| nounCommon | 1,978 | predet | 1 |

Table 5: Lexical Entries in both lexica that did not unify

As explained before, for the cases in table 5 where, although having the same lemma, the entries did not unify the system creates a new entry. This step might cause some undesirable results. This is the case of *no*, encoded as negative adverb in FreeLing with a special tag, where in Apertium it is encoded as a normal adverb. The system creates a new entry, and therefore a duplication. These cases can be traced back when inspecting the log information. The most numerous cases, common nouns and adjectives, mostly correspond to the case of nouns that can also be adjectives, for instance *accesorio* ('incidental' when adjective and 'accessory' when noun). In that case unification fails because of the different PoS value. The system creates a new entry in the resulting lexica, in that case correctly.

## 4. Dicussion

From the results presented above, we can see that using graph unification as merging technique is a successful approach. This method combines compatible information and detects incompatible one, allowing us to keep track of possible merging errors.

Furthermore, the results showed that the technique proposed by Bel et al. (2011) to automatically learn a mapping between lexica that originally encoded information in different ways, have a very good performance in this task. The algorithm correctly learned mapping rules between most of the elements, including those that imply a change in the structure or those that have very few examples in one of the lexica.

In this work we have focused in the use of LMF lexica to test the merging technique, which eases the conversion to feature structures. Though the use of other formats or the conversion to such formats to LMF is an interesting line to be studied in the future, in our opinion the use of LMF is very interesting for different reasons: first of all, because it is a standard format and secondly because it allows the encoding of very complex structures and the possible relations among them. If such structures are encoded in LMF, it is still possible to convert them to feature structures and to perform the automatic mapping and merging, but if these structures are encoded in other formats, discovering them automatically and converting them to a common format with a blind process is much more difficult.

As with respect with previous work, one difference between the application of this technique to SCFs lexica and to morphological lexica is that in the first case, the feature structures obtained after applying the automatic mapping were often incomplete in the sense that some parts of the SCF were partially translated to feature structures and some information was lost. This was overcome in most of the cases at unification step, where the missing information was obtained by subsumption from the target lexicon. Nevertheless, this is not the case in the experiments presented here. In this case, most of the feature structures obtained after applying the mapping are complete and keep all information encoded in the original lexicon. This is partly due to the fact that morphological dictionaries are probably more systematic than SCF lexica, where the SCFs assigned to each verb often have an important variability among lexica. Nevertheless, the improvement observed in the task of merging morphological lexica is also associated to the fact of working with LMF lexica, which allows us to perform a more systematic conversion to feature structures and eases the step of comparing elements of the two lexica. Thus, we can conclude that working with LMF lexica leads to a better performance of our algorithm.

The evaluation we presented here is only qualitative. A proper quantitative intrinsic evaluation will be done by manual inspection as no gold-standard is available. It is also pending to evaluate more accurately the obtained results in an extrinsic evaluation, that is, by assessing the consequences of the errors in another task for comparison.

Summarizing, we have presented an adaptation of the merging mechanism proposed by Bel et al. (2011) to work with LMF lexica that performs fully automatically all steps involved in the merging of two lexica. Furthermore, we have generalized this model to deal with open valued features. This issue was not tackled in the previous work, but is crucial to apply the method to different kind of lexica where this kind of features will be found. The obtained results showed the feasibility of the approach and confirm that this technique can be successfully applied to different kind of lexica.

## 5. Conclusions and Further Work

In this work we have applied the method for automatically merging lexical resources proposed by Bel et al. (2011) to the case of merging morphological dictionaries. These dictionaries were encoded in LMF format, so first of all we adapted the method to work with this standard format and generalized it to deal with open valued features.

The presented experiments showed, on the one hand, that using this method to automatically map two lexica into a common format and then merge them using graph unification mechanism performed satisfactorily in the task tackled in this work. This allows us to make an important step forward to demonstrate the generality of this approach, since it lead to satisfactory results in two very different scenarios: the merging of SCF lexica and the merging of morphological dictionaries.

On the other hand, we have also shown that using LMF as the source encoding format eases the merging process and probably contributes to a better performance of the system.

For that reason, one interesting future line is to study the feasibility of using an approach similar to the mapping technique presented here to convert lexica in any format into LMF. This will help lexicon developers to have their lexica in a standard format.

Another line to be studied in the future is the development of patch rules to refine the obtained results. These patch rules would be the only part of the method dependent on the concrete lexica to be merged and will be developed after systematically detecting possible errors. Besides, we also expect that in order to maximize the usability of the resulting lexica, some scripts can tune the richer and larger lexicon achieved by automatic merging to the requirements of a particular application.

Finally, we are also interested in testing the method for merging other LMF lexica with more or different information (e.g. containing sense information) and especially to apply the proposed technique to the merging of lexica with different levels of information, for example combining the morphological dictionaries with SCF information to obtain a richer, multi-level lexicon.

## 6. Acknowledgements

This work was funded by the EU 7FP project 248064 PANACEA and has greatly benefited of the results of the work done by Marta Villegas in METANET4U project.